\documentclass{article}

    \usepackage[final]{neurips_2021}

\usepackage[utf8]{inputenc} 
\usepackage[T1]{fontenc}    
\usepackage{hyperref}       
\usepackage{url}            
\usepackage{booktabs}       
\usepackage{amsfonts}       
\usepackage{nicefrac}       
\usepackage{microtype}      
\usepackage{xcolor}        
\usepackage{graphicx}
\usepackage{wrapfig}

\title{Bringing Atomistic Deep Learning to Prime Time}

\author{%
  Nathan C. Frey\\
  MIT\\
  \texttt{ncfrey@mit.edu} \\
  \And
    Siddharth Samsi\\
  MIT\\
  \texttt{ssamsi@mit.edu} \\
    \And
    Bharath Ramsundar\\
  Deep Forest Sciences\\
  \texttt{bharath@deepforestsci.com} \\
  \And
    Connor W. Coley\\
  MIT\\
  \texttt{ccoley@mit.edu} \\
    \And
    Vijay Gadepally
    \thanks{DISTRIBUTION STATEMENT A. Approved for public release. Distribution is unlimited. This material is based upon work supported by the Under Secretary of Defense for Research and Engineering under Air Force Contract No. FA8702-15-D-0001. Any opinions, findings, conclusions or recommendations expressed in this material are those of the author(s) and do not necessarily reflect the views of the Under Secretary of Defense for Research and Engineering. © 2021 Massachusetts Institute of Technology. Delivered to the U.S. Government with Unlimited Rights, as defined in DFARS Part 252.227-7013 or 7014 (Feb 2014). Notwithstanding any copyright notice, U.S. Government rights in this work are defined by DFARS 252.227-7013 or DFARS 252.227-7014 as detailed above. Use of this work other than as specifically authorized by the U.S. Government may violate any copyrights that exist in this work.} \\
  MIT\\
  \texttt{vijayg@mit.edu} \\
}

\begin{document}

\maketitle

\begin{abstract}
  Artificial intelligence has not yet revolutionized the design of materials and molecules. In this perspective, we identify four barriers preventing the integration of atomistic deep learning, molecular science, and high-performance computing. We outline focused research efforts to address the opportunities presented by these challenges.
\end{abstract}

\section{Bottlenecks}
\begin{wrapfigure}{R}{0.5\textwidth}
  \begin{center}
  \includegraphics[width=0.48\textwidth]{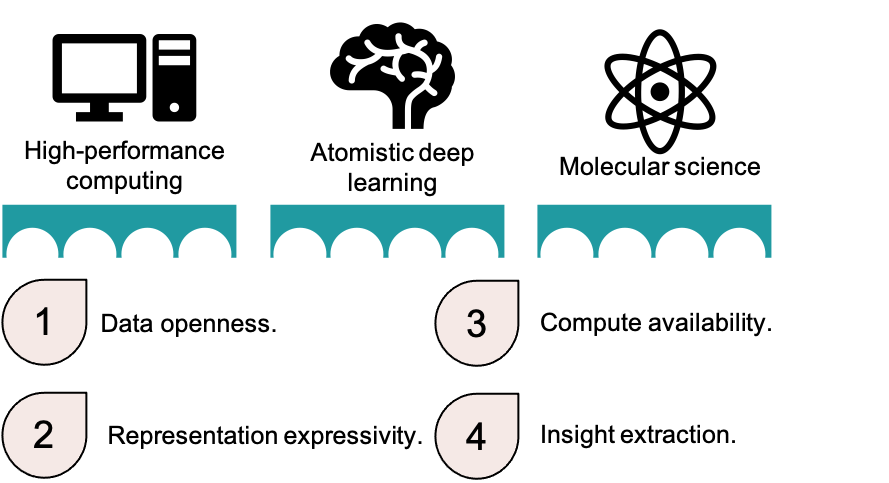}
  \label{fig1}
  \end{center}
  \caption{Key challenges for bridging high-performance computing, atomistic deep learning, and molecular science.}
\end{wrapfigure}

Atomistic deep learning encompasses neural networks that learn representations of matter - from 0D molecules to 1D nanowires, 2D surfaces, 3D proteins, polymers, and crystalline materials [\citenum{Butler2018}]. Unlike other canonical application areas of DL like speech and images, data for atomistic systems is expensive to acquire and heterogeneous, with a practically infinite space of emergent properties of interest. AI revolutionized natural language processing and computer vision, but we are still largely waiting for AI to revolutionize the design of matter. Atomistic systems underpin every physical and digital technology, and the potential impact of AI-enabled matter design cannot be overstated. In this perspective, we identify four key challenges (Figure \ref{fig1}) to the widespread adoption of atomistic DL and outline promising paths forward.

 
The first three challenges are practical difficulties related to robustness and scaling methods to be "production ready." They correspond to the three resources that DL methods traditionally leverage at scale to achieve superhuman success: data, model size, and compute. The fourth challenge is a more fundamental issue related to AI's place in the toolkit for natural scientists.

\paragraph{Data openness}
In many application areas of atomistic DL, practitioners are data starved. Data remains proprietary to the companies that generate it, or not easily accessible to research groups that did not generate the data. Physical experiments and high-fidelity computational simulations can be expensive and time-consuming, so acquiring more labeled data is often not practical. 

There are a number of promising trends addressing this challenge. Adoption of FAIR (findable, accessible, interoperable, reusable) data principles within the research community [\citenum{draxl2018nomad, Andersen_2021}] makes it easier for researchers to leverage data generated outside their own labs. Databases like the Materials Project [\citenum{Jain2013}] and automated experimental validation promise to alleviate the data bottleneck in many domains [\citenum{agrawal2016, himanen2019, Coley2019}]. Active learning and Bayesian optimization strategies for intelligent sampling are shifting the paradigm from random or brute-force data collection to targeted data acquisition [\citenum{Graff_2021, H_se_2021}]. However, it remains important to continue unfocused, high-throughput approaches that generate datasets to train general purpose models and to explore new applications in an open-ended fashion. Datasets generated through intentional experimental design strategies may only be useful to the groups that generated them (and that have the capabilities to generate them).

\vspace{-5pt}
\paragraph{Representation expressivity}
Geometric priors and therefore geometric DL are a natural choice for atomistic systems [\citenum{Kearnes_2016, bronstein2017}], which can often be represented as graphs of atoms (nodes) and chemical bonds (edges). However, graph neural networks often struggle to compete against human-crafted featurizations like chemical fingerprint vectors and much simpler ML methods like random forests and logistic regression in low-data limits [\citenum{mayr2018, Dunn_2020}]. To achieve widespread adoption, atomistic DL methods and molecular representations must be expressive enough to capture the truly staggering diversity of emergent behavior exhibited by matter [\citenum{ramsundar2015massively}], while also being robust and easy to use for non-DL experts.

Following the surprising performance of GPT-3 [\citenum{brown2020language}], one path forward is to leverage large language models trained on string-based representations of matter [\citenum{chithrananda2020chemberta}]. A natural extension is to combine the self-attention mechanism of transformers with the equivariance inherent to molecular graphs [\citenum{fuchs2020se3transformers}]. Identifying optimal pre-training tasks is also an ongoing and fruitful area of research [\citenum{chithrananda2020chemberta, pappu2020making}]. Scaling studies [\citenum{henighan2020scaling}] are needed to investigate whether these architectures and representations are already sufficient to handle generic molecular ML tasks, or if further methodological advances are needed.   

\vspace{-5pt}
\paragraph{Compute availability}
A common thread in headline-grabbing AI achievements is the need for massive compute resources, which are often only available to large corporations. Recent reviews of large language models have argued that non-profit research organizations should also be investigating models at scale [\citenum{bommasani2021opportunities}]. Most atomistic DL methods are not tested beyond small benchmark datasets on a single GPU [\citenum{pappu2020making}]. Even doing a thorough hyperparameter optimization for a moderately-sized dataset can be a burden. As the business value of atomistic DL grows, large tech companies will become increasingly active in the field. To take two recent examples, industry AI research groups have made exciting contributions to catalyst discovery with the Open Catalyst Project [\citenum{zitnick2020introduction}] and protein folding with AlphaFold [\citenum{Jumper2021}]. There is an opportunity to anticipate this trend in atomistic DL and prepare for it, to avoid the paradigm seen in fields like NLP, where fundamental methodological advances are largely left to university research groups and large-scale studies are done by companies.

An emphasis on open science [\citenum{Woelfle2011}] and closer integration between academic research groups and high-performance computing centers, cloud providers, and industry AI groups can alleviate this blocker. Engineering advances like PyTorch Distributed [\citenum{li2020pytorch}] lower the barriers for operating at scale, and HPC centers like MIT Supercloud [\citenum{reuther2018interactive}] and NERSC are increasingly prioritizing GPU resources, GPU acceleration, and interactive ML workflows. As scalable model training becomes easier [\citenum{frey2021scalable}], researchers have the opportunity to investigate emergent behavior in larger models and critically examine where new methods are needed and where scaling existing methods is sufficient.

\vspace{-5pt}
\paragraph{Insight extraction}
Scientists and engineers need to \emph{trust} atomistic DL methods to drive further adoption. Human practitioners don't have an intuition about the failure modes and so there is a resistance to conduct physical experiments based solely on model predictions. Trust is built at a practical level when methods are useful, and at a more basic level when methods provide new, surprising, and relevant scientific insights. Black-box models are useful when they automate a tedious task [\citenum{GOMEZBOMBARELLI20181189}] and the bounds of the model's applicability are well-understood. A loftier goal is for atomistic DL to provide novel insights - discovering new functional forms of matter, uncovering causal or mechanistic processes underpinning empirical observations, etc.

Atomistic DL that discovers not only unexplored, but also synthetically accessible [\citenum{gao2020synthesizability, Frey2019, rgb2016, seifrid2021}], matter will build trust outside the core community of DL-practitioners. Uncertainty quantification of model predictions adds another necessary dimension that will clarify when models are operating outside of their domain of applicability, and uncertainty is a familiar metric for experimentalists to use to assess the trustworthiness of new methods [\citenum{H_se_2021, forde2019scientific}].

\vspace{-5pt}
\paragraph{Conclusion}
The increasing availability of molecular modeling data and the heterogeneity of datasets and tasks present an exciting opportunity for real-world impact by integrating atomistic DL, high-performance computing, and molecular science. Investigating the basic structure of atomistic DL models and their scaling behavior will accelerate the widespread adoption of these methods and uncover what, if anything, learned representations of matter can tell us about matter itself.



\setcitestyle{numbers}
\bibliographystyle{unsrtnat} 
\bibliography{bib}

\begin{thebibliography}{32}
\providecommand{\natexlab}[1]{#1}
\providecommand{\url}[1]{\texttt{#1}}
\expandafter\ifx\csname urlstyle\endcsname\relax
  \providecommand{\doi}[1]{doi: #1}\else
  \providecommand{\doi}{doi: \begingroup \urlstyle{rm}\Url}\fi

\bibitem[Butler et~al.(2018)Butler, Davies, Cartwright, Isayev, and
  Walsh]{Butler2018}
Keith~T. Butler, Daniel~W. Davies, Hugh Cartwright, Olexandr Isayev, and Aron
  Walsh.
\newblock Machine learning for molecular and materials science.
\newblock \emph{Nature}, 559, 2018.
\newblock ISSN 14764687.
\newblock \doi{10.1038/s41586-018-0337-2}.

\bibitem[Draxl and Scheffler(2018)]{draxl2018nomad}
Claudia Draxl and Matthias Scheffler.
\newblock Nomad: The fair concept for big-data-driven materials science, 2018.

\bibitem[Andersen et~al.(2021)Andersen, Armiento, Blokhin, Conduit, Dwaraknath,
  Evans, Fekete, Gopakumar, Gražulis, Merkys, and et~al.]{Andersen_2021}
Casper~W. Andersen, Rickard Armiento, Evgeny Blokhin, Gareth~J. Conduit, Shyam
  Dwaraknath, Matthew~L. Evans, Ádám Fekete, Abhijith Gopakumar, Saulius
  Gražulis, Andrius Merkys, and et~al.
\newblock Optimade, an api for exchanging materials data.
\newblock \emph{Scientific Data}, 8\penalty0 (1), Aug 2021.
\newblock ISSN 2052-4463.
\newblock \doi{10.1038/s41597-021-00974-z}.
\newblock URL \url{http://dx.doi.org/10.1038/s41597-021-00974-z}.

\bibitem[Jain et~al.(2013)Jain, Ong, Hautier, Chen, Richards, Dacek, Cholia,
  Gunter, Skinner, Ceder, and Persson]{Jain2013}
Anubhav Jain, Shyue~Ping Ong, Geoffroy Hautier, Wei Chen, William~Davidson
  Richards, Stephen Dacek, Shreyas Cholia, Dan Gunter, David Skinner, Gerbrand
  Ceder, and Kristin~a. Persson.
\newblock {The Materials Project: A materials genome approach to accelerating
  materials innovation}.
\newblock \emph{APL Materials}, 1\penalty0 (1):\penalty0 011002, 2013.
\newblock ISSN 2166532X.
\newblock \doi{10.1063/1.4812323}.
\newblock URL \url{http://link.aip.org/link/AMPADS/v1/i1/p011002/s1\&Agg=doi}.

\bibitem[Agrawal and Choudhary(2016)]{agrawal2016}
Ankit Agrawal and Alok Choudhary.
\newblock Perspective: Materials informatics and big data: Realization of the
  “fourth paradigm” of science in materials science.
\newblock \emph{APL Materials}, 4\penalty0 (5):\penalty0 053208, 2016.
\newblock \doi{10.1063/1.4946894}.
\newblock URL \url{https://aip.scitation.org/doi/abs/10.1063/1.4946894}.

\bibitem[Himanen et~al.(2019)Himanen, Geurts, Foster, and Rinke]{himanen2019}
Lauri Himanen, Amber Geurts, Adam~Stuart Foster, and Patrick Rinke.
\newblock Data-driven materials science: Status, challenges, and perspectives.
\newblock \emph{Advanced Science}, 6\penalty0 (21):\penalty0 1900808, 2019.
\newblock \doi{https://doi.org/10.1002/advs.201900808}.
\newblock URL
  \url{https://onlinelibrary.wiley.com/doi/abs/10.1002/advs.201900808}.

\bibitem[Coley et~al.(2019)Coley, Thomas, Lummiss, Jaworski, Breen, Schultz,
  Hart, Fishman, Rogers, Gao, Hicklin, Plehiers, Byington, Piotti, Green, Hart,
  Jamison, and Jensen]{Coley2019}
Connor~W. Coley, Dale~A. Thomas, Justin~A.M. Lummiss, Jonathan~N. Jaworski,
  Christopher~P. Breen, Victor Schultz, Travis Hart, Joshua~S. Fishman, Luke
  Rogers, Hanyu Gao, Robert~W. Hicklin, Pieter~P. Plehiers, Joshua Byington,
  John~S. Piotti, William~H. Green, A.~John Hart, Timothy~F. Jamison, and
  Klavs~F. Jensen.
\newblock A robotic platform for flow synthesis of organic compounds informed
  by ai planning.
\newblock \emph{Science}, 365, 2019.
\newblock ISSN 10959203.
\newblock \doi{10.1126/science.aax1566}.

\bibitem[Graff et~al.(2021)Graff, Shakhnovich, and Coley]{Graff_2021}
David~E. Graff, Eugene~I. Shakhnovich, and Connor~W. Coley.
\newblock Accelerating high-throughput virtual screening through molecular
  pool-based active learning.
\newblock \emph{Chemical Science}, 12\penalty0 (22):\penalty0 7866–7881,
  2021.
\newblock ISSN 2041-6539.
\newblock \doi{10.1039/d0sc06805e}.
\newblock URL \url{http://dx.doi.org/10.1039/d0sc06805e}.

\bibitem[Häse et~al.(2021)Häse, Aldeghi, Hickman, Roch, and
  Aspuru-Guzik]{H_se_2021}
Florian Häse, Matteo Aldeghi, Riley~J. Hickman, Loïc~M. Roch, and Alán
  Aspuru-Guzik.
\newblock Gryffin: An algorithm for bayesian optimization of categorical
  variables informed by expert knowledge.
\newblock \emph{Applied Physics Reviews}, 8\penalty0 (3):\penalty0 031406, Sep
  2021.
\newblock ISSN 1931-9401.
\newblock \doi{10.1063/5.0048164}.
\newblock URL \url{http://dx.doi.org/10.1063/5.0048164}.

\bibitem[Kearnes et~al.(2016)Kearnes, McCloskey, Berndl, Pande, and
  Riley]{Kearnes_2016}
Steven Kearnes, Kevin McCloskey, Marc Berndl, Vijay Pande, and Patrick Riley.
\newblock Molecular graph convolutions: moving beyond fingerprints.
\newblock \emph{Journal of Computer-Aided Molecular Design}, 30\penalty0
  (8):\penalty0 595–608, Aug 2016.
\newblock ISSN 1573-4951.
\newblock \doi{10.1007/s10822-016-9938-8}.
\newblock URL \url{http://dx.doi.org/10.1007/s10822-016-9938-8}.

\bibitem[Bronstein et~al.(2017)Bronstein, Bruna, LeCun, Szlam, and
  Vandergheynst]{bronstein2017}
Michael~M. Bronstein, Joan Bruna, Yann LeCun, Arthur Szlam, and Pierre
  Vandergheynst.
\newblock Geometric deep learning: Going beyond euclidean data.
\newblock \emph{IEEE Signal Processing Magazine}, 34\penalty0 (4):\penalty0
  18--42, 2017.
\newblock \doi{10.1109/MSP.2017.2693418}.

\bibitem[Mayr et~al.(2018)Mayr, Klambauer, Unterthiner, Steijaert, Wegner,
  Ceulemans, Clevert, and Hochreiter]{mayr2018}
Andreas Mayr, Günter Klambauer, Thomas Unterthiner, Marvin Steijaert, Jörg~K.
  Wegner, Hugo Ceulemans, Djork-Arné Clevert, and Sepp Hochreiter.
\newblock Large-scale comparison of machine learning methods for drug target
  prediction on chembl.
\newblock \emph{Chem. Sci.}, 9:\penalty0 5441--5451, 2018.
\newblock \doi{10.1039/C8SC00148K}.
\newblock URL \url{http://dx.doi.org/10.1039/C8SC00148K}.

\bibitem[Dunn et~al.(2020)Dunn, Wang, Ganose, Dopp, and Jain]{Dunn_2020}
Alexander Dunn, Qi~Wang, Alex Ganose, Daniel Dopp, and Anubhav Jain.
\newblock Benchmarking materials property prediction methods: the matbench test
  set and automatminer reference algorithm.
\newblock \emph{npj Computational Materials}, 6\penalty0 (1), Sep 2020.
\newblock ISSN 2057-3960.
\newblock \doi{10.1038/s41524-020-00406-3}.
\newblock URL \url{http://dx.doi.org/10.1038/s41524-020-00406-3}.

\bibitem[Ramsundar et~al.(2015)Ramsundar, Kearnes, Riley, Webster, Konerding,
  and Pande]{ramsundar2015massively}
Bharath Ramsundar, Steven Kearnes, Patrick Riley, Dale Webster, David
  Konerding, and Vijay Pande.
\newblock Massively multitask networks for drug discovery, 2015.

\bibitem[Brown et~al.(2020)Brown, Mann, Ryder, Subbiah, Kaplan, Dhariwal,
  Neelakantan, Shyam, Sastry, Askell, Agarwal, Herbert-Voss, Krueger, Henighan,
  Child, Ramesh, Ziegler, Wu, Winter, Hesse, Chen, Sigler, Litwin, Gray, Chess,
  Clark, Berner, McCandlish, Radford, Sutskever, and Amodei]{brown2020language}
Tom~B. Brown, Benjamin Mann, Nick Ryder, Melanie Subbiah, Jared Kaplan,
  Prafulla Dhariwal, Arvind Neelakantan, Pranav Shyam, Girish Sastry, Amanda
  Askell, Sandhini Agarwal, Ariel Herbert-Voss, Gretchen Krueger, Tom Henighan,
  Rewon Child, Aditya Ramesh, Daniel~M. Ziegler, Jeffrey Wu, Clemens Winter,
  Christopher Hesse, Mark Chen, Eric Sigler, Mateusz Litwin, Scott Gray,
  Benjamin Chess, Jack Clark, Christopher Berner, Sam McCandlish, Alec Radford,
  Ilya Sutskever, and Dario Amodei.
\newblock Language models are few-shot learners, 2020.

\bibitem[Chithrananda et~al.(2020)Chithrananda, Grand, and
  Ramsundar]{chithrananda2020chemberta}
Seyone Chithrananda, Gabriel Grand, and Bharath Ramsundar.
\newblock Chemberta: Large-scale self-supervised pretraining for molecular
  property prediction, 2020.

\bibitem[Fuchs et~al.(2020)Fuchs, Worrall, Fischer, and
  Welling]{fuchs2020se3transformers}
Fabian~B. Fuchs, Daniel~E. Worrall, Volker Fischer, and Max Welling.
\newblock Se(3)-transformers: 3d roto-translation equivariant attention
  networks, 2020.

\bibitem[Pappu and Paige(2020)]{pappu2020making}
Aneesh Pappu and Brooks Paige.
\newblock Making graph neural networks worth it for low-data molecular machine
  learning, 2020.

\bibitem[Henighan et~al.(2020)Henighan, Kaplan, Katz, Chen, Hesse, Jackson,
  Jun, Brown, Dhariwal, Gray, Hallacy, Mann, Radford, Ramesh, Ryder, Ziegler,
  Schulman, Amodei, and McCandlish]{henighan2020scaling}
Tom Henighan, Jared Kaplan, Mor Katz, Mark Chen, Christopher Hesse, Jacob
  Jackson, Heewoo Jun, Tom~B. Brown, Prafulla Dhariwal, Scott Gray, Chris
  Hallacy, Benjamin Mann, Alec Radford, Aditya Ramesh, Nick Ryder, Daniel~M.
  Ziegler, John Schulman, Dario Amodei, and Sam McCandlish.
\newblock Scaling laws for autoregressive generative modeling, 2020.

\bibitem[Bommasani et~al.(2021)Bommasani, Hudson, Adeli, Altman, Arora, von
  Arx, Bernstein, Bohg, Bosselut, Brunskill, Brynjolfsson, Buch, Card,
  Castellon, Chatterji, Chen, Creel, Davis, Demszky, Donahue, Doumbouya,
  Durmus, Ermon, Etchemendy, Ethayarajh, Fei-Fei, Finn, Gale, Gillespie, Goel,
  Goodman, Grossman, Guha, Hashimoto, Henderson, Hewitt, Ho, Hong, Hsu, Huang,
  Icard, Jain, Jurafsky, Kalluri, Karamcheti, Keeling, Khani, Khattab, Kohd,
  Krass, Krishna, Kuditipudi, Kumar, Ladhak, Lee, Lee, Leskovec, Levent, Li,
  Li, Ma, Malik, Manning, Mirchandani, Mitchell, Munyikwa, Nair, Narayan,
  Narayanan, Newman, Nie, Niebles, Nilforoshan, Nyarko, Ogut, Orr,
  Papadimitriou, Park, Piech, Portelance, Potts, Raghunathan, Reich, Ren, Rong,
  Roohani, Ruiz, Ryan, Ré, Sadigh, Sagawa, Santhanam, Shih, Srinivasan,
  Tamkin, Taori, Thomas, Tramèr, Wang, Wang, Wu, Wu, Wu, Xie, Yasunaga, You,
  Zaharia, Zhang, Zhang, Zhang, Zhang, Zheng, Zhou, and
  Liang]{bommasani2021opportunities}
Rishi Bommasani, Drew~A. Hudson, Ehsan Adeli, Russ Altman, Simran Arora, Sydney
  von Arx, Michael~S. Bernstein, Jeannette Bohg, Antoine Bosselut, Emma
  Brunskill, Erik Brynjolfsson, Shyamal Buch, Dallas Card, Rodrigo Castellon,
  Niladri Chatterji, Annie Chen, Kathleen Creel, Jared~Quincy Davis, Dora
  Demszky, Chris Donahue, Moussa Doumbouya, Esin Durmus, Stefano Ermon, John
  Etchemendy, Kawin Ethayarajh, Li~Fei-Fei, Chelsea Finn, Trevor Gale, Lauren
  Gillespie, Karan Goel, Noah Goodman, Shelby Grossman, Neel Guha, Tatsunori
  Hashimoto, Peter Henderson, John Hewitt, Daniel~E. Ho, Jenny Hong, Kyle Hsu,
  Jing Huang, Thomas Icard, Saahil Jain, Dan Jurafsky, Pratyusha Kalluri,
  Siddharth Karamcheti, Geoff Keeling, Fereshte Khani, Omar Khattab, Pang~Wei
  Kohd, Mark Krass, Ranjay Krishna, Rohith Kuditipudi, Ananya Kumar, Faisal
  Ladhak, Mina Lee, Tony Lee, Jure Leskovec, Isabelle Levent, Xiang~Lisa Li,
  Xuechen Li, Tengyu Ma, Ali Malik, Christopher~D. Manning, Suvir Mirchandani,
  Eric Mitchell, Zanele Munyikwa, Suraj Nair, Avanika Narayan, Deepak
  Narayanan, Ben Newman, Allen Nie, Juan~Carlos Niebles, Hamed Nilforoshan,
  Julian Nyarko, Giray Ogut, Laurel Orr, Isabel Papadimitriou, Joon~Sung Park,
  Chris Piech, Eva Portelance, Christopher Potts, Aditi Raghunathan, Rob Reich,
  Hongyu Ren, Frieda Rong, Yusuf Roohani, Camilo Ruiz, Jack Ryan, Christopher
  Ré, Dorsa Sadigh, Shiori Sagawa, Keshav Santhanam, Andy Shih, Krishnan
  Srinivasan, Alex Tamkin, Rohan Taori, Armin~W. Thomas, Florian Tramèr,
  Rose~E. Wang, William Wang, Bohan Wu, Jiajun Wu, Yuhuai Wu, Sang~Michael Xie,
  Michihiro Yasunaga, Jiaxuan You, Matei Zaharia, Michael Zhang, Tianyi Zhang,
  Xikun Zhang, Yuhui Zhang, Lucia Zheng, Kaitlyn Zhou, and Percy Liang.
\newblock On the opportunities and risks of foundation models, 2021.

\bibitem[Zitnick et~al.(2020)Zitnick, Chanussot, Das, Goyal, Heras-Domingo, Ho,
  Hu, Lavril, Palizhati, Riviere, Shuaibi, Sriram, Tran, Wood, Yoon, Parikh,
  and Ulissi]{zitnick2020introduction}
C.~Lawrence Zitnick, Lowik Chanussot, Abhishek Das, Siddharth Goyal, Javier
  Heras-Domingo, Caleb Ho, Weihua Hu, Thibaut Lavril, Aini Palizhati, Morgane
  Riviere, Muhammed Shuaibi, Anuroop Sriram, Kevin Tran, Brandon Wood, Junwoong
  Yoon, Devi Parikh, and Zachary Ulissi.
\newblock An introduction to electrocatalyst design using machine learning for
  renewable energy storage, 2020.

\bibitem[Jumper et~al.(2021)Jumper, Evans, Pritzel, Green, Figurnov,
  Ronneberger, Tunyasuvunakool, Bates, Žídek, Potapenko, Bridgland, Meyer,
  Kohl, Ballard, Cowie, Romera-Paredes, Nikolov, Jain, Adler, Back, Petersen,
  Reiman, Clancy, Zielinski, Steinegger, Pacholska, Berghammer, Bodenstein,
  Silver, Vinyals, Senior, Kavukcuoglu, Kohli, and Hassabis]{Jumper2021}
John Jumper, Richard Evans, Alexander Pritzel, Tim Green, Michael Figurnov,
  Olaf Ronneberger, Kathryn Tunyasuvunakool, Russ Bates, Augustin Žídek, Anna
  Potapenko, Alex Bridgland, Clemens Meyer, Simon~A.A. Kohl, Andrew~J. Ballard,
  Andrew Cowie, Bernardino Romera-Paredes, Stanislav Nikolov, Rishub Jain,
  Jonas Adler, Trevor Back, Stig Petersen, David Reiman, Ellen Clancy, Michal
  Zielinski, Martin Steinegger, Michalina Pacholska, Tamas Berghammer,
  Sebastian Bodenstein, David Silver, Oriol Vinyals, Andrew~W. Senior, Koray
  Kavukcuoglu, Pushmeet Kohli, and Demis Hassabis.
\newblock Highly accurate protein structure prediction with alphafold.
\newblock \emph{Nature}, 2021.
\newblock ISSN 14764687.
\newblock \doi{10.1038/s41586-021-03819-2}.

\bibitem[Woelfle et~al.(2011)Woelfle, Olliaro, and Todd]{Woelfle2011}
Michael Woelfle, Piero Olliaro, and Matthew~H. Todd.
\newblock Open science is a research accelerator.
\newblock \emph{Nature Chemistry}, 3, 2011.
\newblock ISSN 17554330.
\newblock \doi{10.1038/nchem.1149}.

\bibitem[Li et~al.(2020)Li, Zhao, Varma, Salpekar, Noordhuis, Li, Paszke,
  Smith, Vaughan, Damania, and Chintala]{li2020pytorch}
Shen Li, Yanli Zhao, Rohan Varma, Omkar Salpekar, Pieter Noordhuis, Teng Li,
  Adam Paszke, Jeff Smith, Brian Vaughan, Pritam Damania, and Soumith Chintala.
\newblock Pytorch distributed: Experiences on accelerating data parallel
  training, 2020.

\bibitem[Reuther et~al.(2018)Reuther, Kepner, Byun, Samsi, Arcand, Bestor,
  Bergeron, Gadepally, Houle, Hubbell, Jones, Klein, Milechin, Mullen, Prout,
  Rosa, Yee, and Michaleas]{reuther2018interactive}
Albert Reuther, Jeremy Kepner, Chansup Byun, Siddharth Samsi, William Arcand,
  David Bestor, Bill Bergeron, Vijay Gadepally, Michael Houle, Matthew Hubbell,
  Michael Jones, Anna Klein, Lauren Milechin, Julia Mullen, Andrew Prout,
  Antonio Rosa, Charles Yee, and Peter Michaleas.
\newblock Interactive supercomputing on 40,000 cores for machine learning and
  data analysis.
\newblock In \emph{2018 IEEE High Performance extreme Computing Conference
  (HPEC)}, pages 1--6. IEEE, 2018.

\bibitem[Frey et~al.(2021)Frey, Samsi, McDonald, Li, Coley, and
  Gadepally]{frey2021scalable}
Nathan~C. Frey, Siddharth Samsi, Joseph McDonald, Lin Li, Connor~W. Coley, and
  Vijay Gadepally.
\newblock Scalable geometric deep learning on molecular graphs, 2021.

\bibitem[Gómez-Bombarelli(2018)]{GOMEZBOMBARELLI20181189}
Rafael Gómez-Bombarelli.
\newblock Reaction: The near future of artificial intelligence in materials
  discovery.
\newblock \emph{Chem}, 4\penalty0 (6):\penalty0 1189--1190, 2018.
\newblock ISSN 2451-9294.
\newblock \doi{https://doi.org/10.1016/j.chempr.2018.05.021}.
\newblock URL
  \url{https://www.sciencedirect.com/science/article/pii/S2451929418302304}.

\bibitem[Gao and Coley(2020)]{gao2020synthesizability}
Wenhao Gao and Connor~W. Coley.
\newblock The synthesizability of molecules proposed by generative models,
  2020.

\bibitem[Frey et~al.(2019)Frey, Wang, Bellido, Anasori, Gogotsi, and
  Shenoy]{Frey2019}
Nathan~C. Frey, Jin Wang, Gabriel Iván~Vega Bellido, Babak Anasori, Yury
  Gogotsi, and Vivek~B. Shenoy.
\newblock Prediction of synthesis of 2d metal carbides and nitrides (mxenes)
  and their precursors with positive and unlabeled machine learning.
\newblock \emph{ACS Nano}, 13:\penalty0 3031--3041, 2019.
\newblock ISSN 1936086X.
\newblock \doi{10.1021/acsnano.8b08014}.
\newblock URL \url{http://pubs.acs.org/doi/10.1021/acsnano.8b08014}.

\bibitem[Gómez-Bombarelli et~al.(2016)Gómez-Bombarelli,
  Aguilera-Iparraguirre, Hirzel, Duvenaud, Maclaurin, Blood-Forsythe, Chae,
  Einzinger, Ha, Wu, Markopoulos, Jeon, Kang, Miyazaki, Numata, Kim, Huang,
  Hong, Baldo, Adams, and Aspuru-Guzik]{rgb2016}
Rafael Gómez-Bombarelli, Jorge Aguilera-Iparraguirre, Timothy~D Hirzel, David
  Duvenaud, Dougal Maclaurin, Martin~A Blood-Forsythe, Hyun~Sik Chae, Markus
  Einzinger, Dong-Gwang Ha, Tony Wu, Georgios Markopoulos, Soonok Jeon, Hosuk
  Kang, Hiroshi Miyazaki, Masaki Numata, Sunghan Kim, Wenliang Huang, Seong~Ik
  Hong, Marc Baldo, Ryan~P Adams, and Alán Aspuru-Guzik.
\newblock Design of efficient molecular organic light-emitting diodes by a
  high-throughput virtual screening and experimental approach.
\newblock \emph{Nature Materials}, 15:\penalty0 1120, 2016.
\newblock \doi{10.1038/nmat4717
  https://www.nature.com/articles/nmat4717#supplementary-information}.
\newblock URL \url{http://dx.doi.org/10.1038/nmat4717}.

\bibitem[Seifrid et~al.(2021)Seifrid, Hickman, Aguilar-Granda, Lavigne,
  Vestfrid, Wu, Gaudin, Hopkins, and Aspuru-Guzik]{seifrid2021}
Martin Seifrid, Riley~J. Hickman, Andrés Aguilar-Granda, Cyrille Lavigne,
  Jenya Vestfrid, Tony~C. Wu, Théophile Gaudin, Emily~J. Hopkins, and Alán
  Aspuru-Guzik.
\newblock Routescore: Punching the ticket to more efficient materials
  development.
\newblock \emph{ChemRxiv}, 2021.
\newblock \doi{10.33774/chemrxiv-2021-k0qx5}.

\bibitem[Forde and Paganini(2019)]{forde2019scientific}
Jessica~Zosa Forde and Michela Paganini.
\newblock The scientific method in the science of machine learning, 2019.

\end{thebibliography}

\end{document}